\documentclass[10pt,twocolumn,letterpaper]{article}

\usepackage[pagenumbers]{cvpr}

\makeatletter
\@namedef{ver@everyshi.sty}{}
\makeatother
\usepackage{tikz}
\usepackage{pgfplots}
\usepackage{pgfplotstable}

\usepackage{graphicx}
\usepackage{amsmath}
\usepackage{amssymb}
\usepackage{booktabs}
\usepackage{float}
\usepackage{array}
\usepackage{adjustbox}
\usepackage{multirow}
\usepackage[symbol]{footmisc}

\usepackage[pagebackref,breaklinks,colorlinks]{hyperref}

\definecolor{spectral1}{HTML}{9E0142}
\definecolor{spectral2}{HTML}{D53E4F}
\definecolor{spectral3}{HTML}{F46D43}
\definecolor{spectral4}{HTML}{FDAE61}
\definecolor{spectral5}{HTML}{606669} 
\definecolor{spectral6}{HTML}{FFFFBF}
\definecolor{spectral7}{HTML}{E6F598}
\definecolor{spectral8}{HTML}{ABDDA4}
\definecolor{spectral9}{HTML}{66C2A5}
\definecolor{spectral10}{HTML}{3288BD}
\definecolor{spectral11}{HTML}{5E4FA2}

\usepackage[capitalize]{cleveref}
\crefname{section}{Sec.}{Secs.}
\Crefname{section}{Section}{Sections}
\Crefname{table}{Table}{Tables}
\crefname{table}{Tab.}{Tabs.}

\renewcommand{\vec}[1]{\boldsymbol{#1}}
\DeclareMathOperator{\loss}{\mathcal{L}}
\DeclareMathOperator{\E}{\mathbb{E}}

\usepackage{enumitem}

\newcommand\blfootnote[1]{%
  \begingroup
  \renewcommand\thefootnote{}\footnote{#1}%
  \addtocounter{footnote}{-1}%
  \endgroup
}

\begin{document}

\title{Unleashing Transformers: Parallel Token Prediction with Discrete Absorbing Diffusion \\for Fast High-Resolution Image Generation from Vector-Quantized Codes}

\author{Sam Bond-Taylor\thanks{Authors contributed equally.}\ $^1$, Peter Hessey$^{*1}$, Hiroshi Sasaki$^1$, Toby P. Breckon$^{1,2}$, Chris G. Willcocks$^1$\\
Department of \{$^1$Computer Science $|$ $^2$Engineering\},
Durham University,
Durham, UK
}
\maketitle

\begin{abstract}
  Whilst diffusion probabilistic models can generate high quality image content, key limitations remain in terms of both generating high-resolution imagery and their associated high computational requirements. Recent Vector-Quantized image models have overcome this limitation of image resolution but are prohibitively slow and unidirectional as they generate tokens via element-wise autoregressive sampling from the prior. By contrast, in this paper we propose a novel discrete diffusion probabilistic model prior which enables parallel prediction of Vector-Quantized tokens by using an unconstrained Transformer architecture as the backbone. During training, tokens are randomly masked in an order-agnostic manner and the Transformer learns to predict the original tokens. This parallelism of Vector-Quantized token prediction in turn facilitates unconditional generation of globally consistent high-resolution and diverse imagery at a fraction of the computational expense. In this manner, we can generate image resolutions exceeding that of the original training set samples whilst additionally provisioning per-image likelihood estimates (in a departure from generative adversarial approaches). Our approach achieves state-of-the-art results in terms of Density (LSUN Bedroom: 1.51; LSUN Churches: 1.12; FFHQ: 1.20) and Coverage (LSUN Bedroom: 0.83; LSUN Churches: 0.73; FFHQ: 0.80), and performs competitively on FID (LSUN Bedroom: 3.64; LSUN Churches: 4.07; FFHQ: 6.11) whilst offering advantages in terms of both computation and reduced training set requirements.
\end{abstract}
\vspace{-1.9em} 

\blfootnote{\hspace{-0.9em}Source code for this work is available at \url{https://github.com/samb-t/unleashing-transformers}}

\section{Introduction}

\begin{figure}[t]
    \centering
    \includegraphics[width=\linewidth]{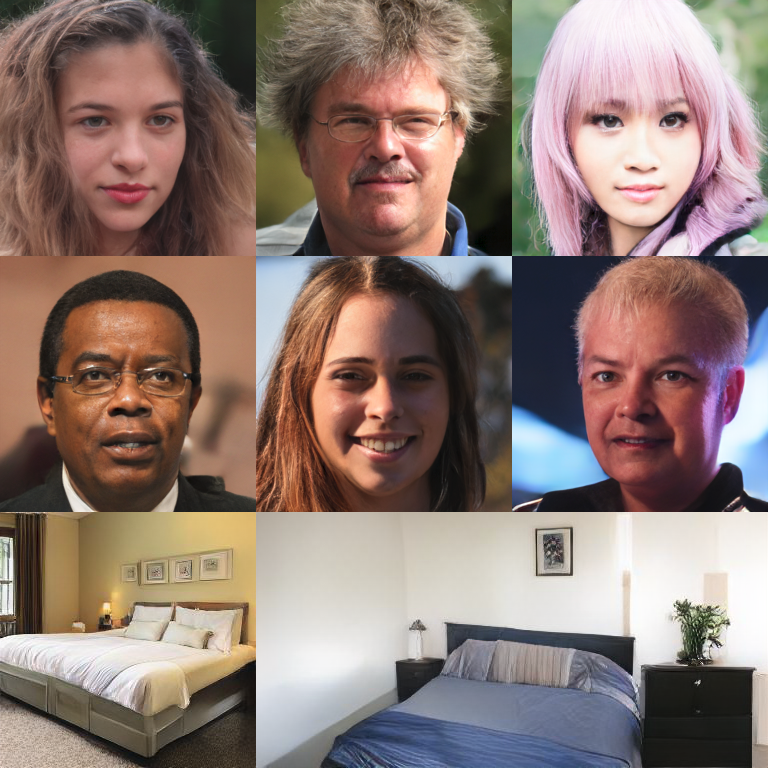}
    \caption{Our approach uses a discrete diffusion to generate high quality images optionally larger than the training data (bottom).}
\end{figure}

\noindent Artificially generating plausible photo-realistic images, at ever higher resolutions, has long been a goal when designing deep generative models. Recent advancements have yielded direct benefits for fields such as medical image synthesis \cite{fetty2020latent}, computer graphics \cite{chen2021towards, yu2019texture}, image editing \cite{lin2021anycost}, image-to-image translation \cite{sasaki2021unit}, and image super-resolution \cite{ho2021cascaded}. 

These methods can in general be divided into five main classes \cite{bond2021deep}, each of which make different trade-offs to scale to high resolutions. Techniques to scale Generative Adversarial Networks (GANs) \cite{goodfellow2014generative} include progressive growing \cite{karras2018progressive}, large batches \cite{brock2019large}, and regularisation \cite{miyato2018spectral, liu2021faster}. Variational Autoencoders (VAEs) \cite{kingma2014autoencoding} can be scaled by building complex priors \cite{child2021very, vahdat2021score, VanDenOord2017NeuralDiscreteRepresentation} and correcting the learned density \cite{xiao2021vaebm}. Autoregressive approaches can make independence assumptions \cite{reed2017parallel} or partition spatial dimensions \cite{menick2019generating}. Normalizing Flows utilise multi-scale architectures \cite{kingma2018glow}, while diffusion models can be scaled using SDEs \cite{song2021scorebased} and cascades \cite{ho2021cascaded}. Each of these approaches have their own drawbacks, such as unstable training, long sample times, and a lack of global context.

Of particular interest to this work is the popular Transformer architecture \cite{vaswani2017attention} which is able to model long distance relationships using a powerful attention mechanism that can be trained in parallel. By constraining the Transformer architecture to attend a fixed ordering of tokens in a unidirectional manner, they can be used to parameterise an autoregressive model for generative modelling \cite{child2019generating, parmar2018image}. However, image data does not conform to such a structure and hence  this bias limits the representation ability of the Transformer and unnecessarily restricts the sampling process to be both sequential and slow. 
\vspace{0.1cm}

\noindent Addressing these issues, our main contributions are:
\begin{itemize}[itemsep=2pt,parsep=0pt,topsep=2pt]
    \item We propose a novel parallel token prediction approach for generating Vector-Quantized image representations that allows for significantly faster sampling than autoregressive approaches.
    \item Our approach is able to generate globally consistent images at resolutions exceeding that of the original training data by aggregating multiple context windows, allowing for much larger context regions.
    \item Our approach demonstrates state-of-the art performance across three benchmark datasets in terms of Density (LSUN Bedroom: 1.51; LSUN Churches: 1.12; FFHQ: 1.20) and Coverage (LSUN Bedroom: 0.83; LSUN Churches: 0.73; FFHQ: 0.80), while also being competitive on FID (LSUN Bedroom: 3.64; LSUN Churches: 4.07; FFHQ: 6.11). 
\end{itemize}

\section{Prior Work}
\vspace{-0.15em}
\noindent Extensive work in deep generative modelling \cite{bond2021deep} and self-supervised learning \cite{ericsson2021well} laid the foundations for this research, which we review here in terms of both existing models (Sections 2.1-2.4) and  the Transformer architecture (Section 2.5).

\subsection{Autoregressive Models}
\vspace{-0.17em}
\noindent Autoregressive models are a family of powerful generative models capable of directly maximising the likelihood of the data on which they are trained. These models have achieved impressive image generation results in recent years, however, their sequential nature limits them to relatively low dimensional data \cite{VanDenOord2016PixelRecurrentNeural, VanDenOord2016ConditionalImageGeneration, Child2019GeneratingLongSequences, jun2020distribution, Salimans2017PixelCNNImprovingPixelCNN, roy2021efficient}. 

The training and inference process for autoregressive models is based on the chain rule of probability. By decomposing inputs into their individual components $\vec{x} = \{x_1, ..., x_n\}$, an autoregressive model with parameters $\theta$ can generate new latent samples sequentially:
\vspace{-0.4em}
\begin{equation}
    p_\theta(\vec{x}) = p_\theta(x_1, x_2, ..., x_n) = \prod_{i=1}^{n}p_\theta(x_i|x_1, ..., x_{i-1}).
    \vspace{-0.2em}
\end{equation}
Selecting an ordering over inputs is not obvious for many tasks; since the receptive field is limited to previously generated tokens, this can significantly affect sample quality.

\subsection{Discrete Energy-Based Models}
\vspace{-0.17em}
\noindent Since the causal nature of autoregressive models limits their representation ability, other approaches with less constrained architectures have begun to outperform them even on likelihood \cite{kingma2021variational}.
Energy-based models (EBMs) are an enticing method for representing discrete data as they permit unconstrained architectures with global context. Implicit EBMs define an unnormalised distribution over data that is typically learned through contrastive divergence \cite{du2019implicit,hinton2002training}. Unfortunately, sampling EBMs using Gibbs sampling is impractical for high dimensional discrete data. However, gradients can be incorporated to reduce mixing times \cite{grathwohl2021oops}.

Similar to autoregressive models, masked language models (MLMs) such as BERT \cite{devlin2019bert} model the conditional probability of the data. However, these are trained bidirectionally by randomly masking a subset of tokens from the input sequence, allowing a much richer context than autoregressive approaches. Some attempts have been made to define an implicit energy function using the conditional probabilities \cite{wang2019bert}, however, obtaining true samples leads to very long sample times and we found them to be ineffective at modelling longer sequences during our experiments \cite{goyal2021exposing}.

\begin{figure*}[t]
    \centering
    \includegraphics[width=0.98\linewidth,trim={0 0 0.74cm 0},clip]{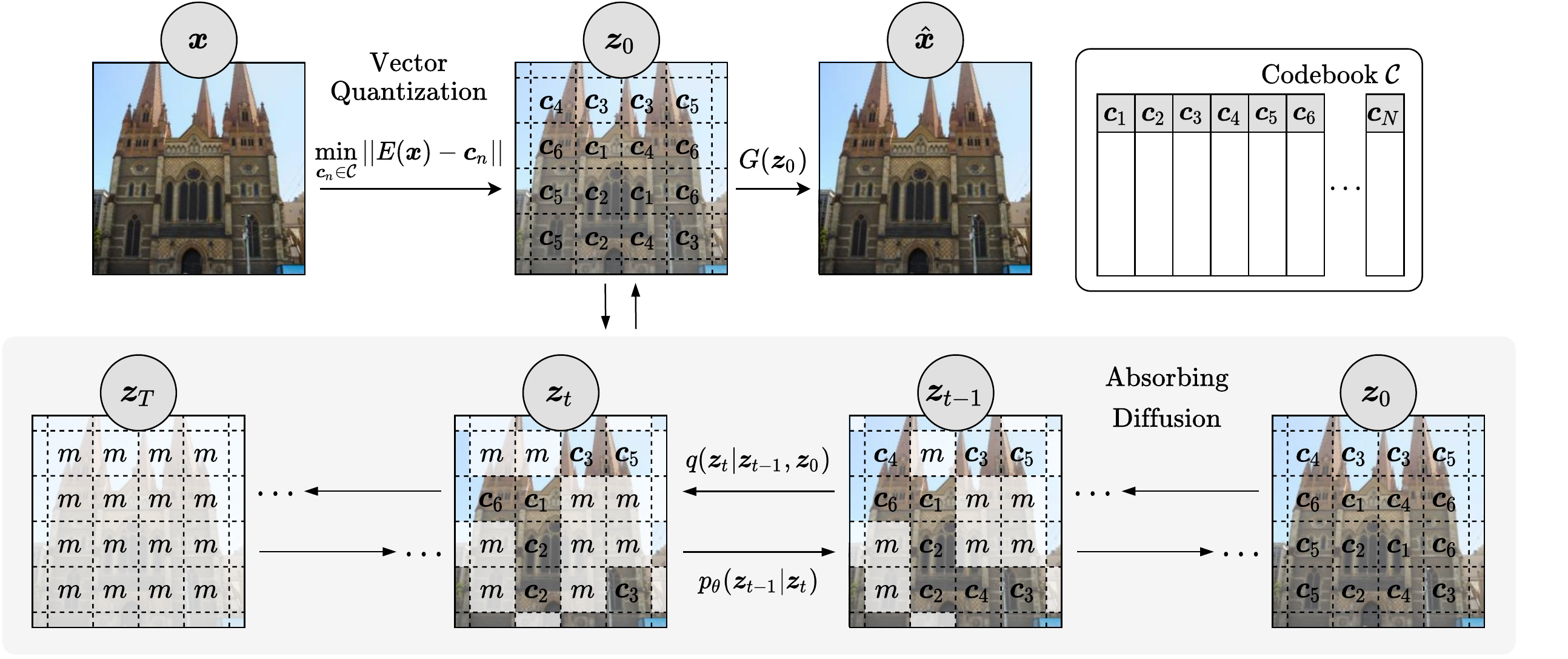}
    \caption{Our approach uses a discrete absorbing diffusion model to represent Vector-Quantized images allowing fast high-resolution image generation. Specifically, after compressing images to an information-rich discrete space, elements are randomly masked and an unconstrained Transformer is trained to predict the original data, using global context to ensure samples are consistent and high quality.}
    \label{fig:main-diagram}
\end{figure*}

\subsection{Discrete Denoising Diffusion Models}
\noindent Diffusion models \cite{sohl-dickstein2015deep, ho2020denoising} define a Markov chain $q(\vec{x}_{1:T}|\vec{x}_0) = \prod_{t=1}^T q(\vec{x}_t| \vec{x}_{t-1})$ that gradually destroys data $\vec{x}_0$ by adding noise over a fixed number of steps $T$ so that $\vec{x}_T$ contains little to no information about $\vec{x}_0$ and can be easily sampled. The reverse procedure is a generative model that gradually denoises towards the data distribution $p_\theta(\vec{x}_{0:T}) = p_\theta(\vec{x}_T) \prod_{t=1}^T p_\theta(\vec{x}_{t-1}|\vec{x}_t)$, learned by optimising the evidence lower bound (ELBO), with $t^\text{th}$ term 
\begin{equation}
    \E_{q(\vec{x}_{t+1}|\vec{x}_0)}\big[ D_{KL}( q(\vec{x}_t|\vec{x}_{t+1}, \vec{x}_0) || p_\theta(\vec{x}_t|\vec{x}_{t+1}) ) \big],
\end{equation}
where sampling from the reverse process is not required during training. When applied in continuous spaces, distributions are typically parameterised as Normal distributions.

Discrete diffusion models \cite{sohl-dickstein2015deep, hoogeboom2021argmax, austin2021structured} constrain the state space so that $\vec{x}_t$ is a discrete random variable falling into one of $K$ categories. As such, the forward process can be represented as categorical distributions $q(\vec{x}_t|\vec{x}_{t-1})= \text{Cat}(\vec{x}_t;\vec{p}=\vec{x}_{t-1} \vec{Q}_t)$ for one-hot $\vec{x}_{t-1}$ where $\vec{Q}_t$ is a matrix denoting the probabilities of moving to each successive state. Transition processes include moving states with some low uniform probability \cite{hoogeboom2021argmax}, moving to nearby states with some low probability based on similarity or distance, and masking out values entirely similar to generative MLMs.

\subsection{Hybrid Generative Models}
\noindent Hybrid models combine two or more classes of generative model to balance trade-offs such as slow sampling, poor scaling with dimension, and inadequate modelling flexibility. Many of these approaches are based on Variational Autoencoders (VAEs) \cite{kingma2014autoencoding, rezende2014stochastic} which have fast run-times and scale well to high resolutions, but struggle with sample quality. For example, a VAE's  approximate posterior and/or prior complexity can be increased by applying a second generative model such as a Normalizing Flow \cite{kingma2016improved,grathwohl2018ffjord,vahdat2020nvae,vahdat2021score} or EBM \cite{pang2020learning} in latent space. Alternatively, a second model can be used to correct samples \cite{xiao2021vaebm}. Of particular interest to this work are Vector-Quantized image models, which follow a 2-stage training scheme where a convolutional autoencoder extracts high level features to an information rich discrete latent space and a powerful autoregressive density estimator learns the prior over these latents \cite{VanDenOord2017NeuralDiscreteRepresentation, esser2021taming,Razavi2019GeneratingDiverseHighFidelity}. 

\subsection{Transformers}
\noindent Transformers \cite{vaswani2017attention} have made a huge impact across many fields of deep learning \cite{han2020survey} due to their power and flexibility. They are based on the concept of self-attention, a function which allows interactions with strong gradients between all inputs, irrespective of their spatial relationships. This procedure (Eqn. \ref{eqn:attention}) encodes inputs as key-value pairs, where values $\vec{V}$ represent embedded inputs and keys $\vec{K}$ act as an indexing method, subsequently, a set of queries $\vec{Q}$ are used to select which values to observe:
\begin{equation}\label{eqn:attention}
    \text{Attn}(\vec{Q}, \vec{K}, \vec{V}) = \text{softmax}\bigg( \frac{\vec{Q}\vec{K}^T}{\sqrt{d_k}} \bigg) \vec{V}.
\end{equation}
While this allows long distance dependencies to be learned, complexity increases with sequence length quadratically, making scaling to high dimensional inputs difficult. Approaches to mitigate this include independence assumptions \cite{reed2017parallel}, sparsity \cite{child2019generating}, and kernel approximations \cite{katharopoulos2020transformers}.
\section{Method}\label{sec:method}
\noindent In this section, we formalise our proposed 2-stage approach for generating high-resolution images using a discrete diffusion model to represent Vector-Quantized image representations; this is visualised in \cref{fig:main-diagram}. We hypothesise that by removing the autoregressive constraint, allowing bidirectional context when generating samples, not only will it be possible to speed up the sampling process, but an improved feature representation will be learned, enabling higher quality image generation. 

\subsection{Learning Codes}
\noindent In the first stage of our approach, a Vector-Quantized image model compresses high-resolution images to a highly compressed form, taking advantage of an information rich codebook \cite{VanDenOord2017NeuralDiscreteRepresentation}. A convolutional encoder downsamples images $\vec{x}$ to a smaller spatial resolution, $E(\vec{x}) = \{\vec{e}_1, \vec{e}_2, ..., \vec{e}_L\} \in \mathbb{R}^{L \times D}$. A simple quantisation approach is to use the $\arg\!\max$ operation which maps continuous encodings to their closest elements in a finite codebook of vectors \cite{VanDenOord2017NeuralDiscreteRepresentation}. Specifically, for a codebook $\mathcal{C} \in \mathbb{R}^{K \times D}$, where $K$ is the number of discrete codes in the codebook and $D$ is the dimension of each code, each $\vec{e}_i$ is mapped via a nearest-neighbour lookup onto a discrete codebook value, $\vec{c}_j \in \mathcal{C}$:
\begin{equation}\label{eqn:quantisation}
    \hspace{-0.05em}\vec{z}_q = \{\vec{q}_1, \vec{q}_2, ..., \vec{q}_L\} \text{  , where  } \vec{q}_i = \underset{\vec{c}_{j} \in \mathcal{C}}{\operatorname{min}}||\vec{e}_i - \vec{c}_j||.
\end{equation}
As this operation is non-differentiable, the straight-through gradient estimator \cite{Bengio2013Estimating} is used to copy the gradients from the decoder inputs onto the encoder outputs resulting in biased gradients. Subsequently, the quantized latents are fed through a decoder network $\hat{\vec{x}}= G(\vec{z}_q)$ to reconstruct the input based on a perceptual reconstruction loss \cite{zhang2018unreasonable, esser2021taming}; this process is trained by minimising the loss $\loss_\text{VQ}$,
\begin{equation}\label{eqn:vqvae-loss}
    \hspace{-0.4em}\loss_\text{VQ} = \loss_\text{rec} + || \text{sg}[E(\vec{x})] - \vec{z}_q ||_2^2 + \beta || \text{sg}[\vec{z}_q] - E(\vec{x}) ||_2^2. \hspace{-0.1em}
\end{equation}
The $\arg\!\max$ approach can result in codebook collapse, where some codes are never used; while other quantisation methods can reduce this \cite{dieleman2018challenge, maddison2017concrete, jang2017categorical, ramesh2021zero}, we found $\arg\!\max$ quantisation to yield the highest reconstruction quality.

\subsection{Sampling Globally Coherent Latents} \label{sec:method-global}
\noindent To allow sampling, a discrete generative model is trained on the latents obtained from the Vector-Quantized image model. The highly compressed form allows this second stage to function much more efficiently. Once the training data is encoded as discrete, integer-valued latents  $\vec{z} \in \mathbb{Z}^D$, a discrete diffusion model can be used to learn the distribution over these latents. Due to the effectiveness of BERT-style models \cite{devlin2019bert} for representation learning, we use the absorbing state diffusion \cite{austin2021structured} which similarly learns to denoise randomly masked data. Specifically, in each forward time step $t$, values are either kept the same or masked out entirely with probability $\frac{1}{t}$ and the reverse process gradually unveils these masks. Rather than directly approximating $p_\theta(\vec{z}_{t-1}|\vec{z}_t)$, we predict $p_\theta(\vec{z}_0|\vec{z}_t)$, reducing the training stochasticity \cite{ho2020denoising}. The variational bound reduces to
\vspace{-0.2em}
\begin{equation}
    \label{eqn:dp3m-absorbing-elbo}
    \hspace*{-0.2em} \E_{q(\vec{z}_0)} \Bigg[ \sum_{t=1}^T \frac{1}{t} \E_{q(\vec{z}_t|\vec{z}_0)} \Big[ \sum_{[\vec{z}_t]_i=m} \log p_\theta([\vec{z}_0]_i|\vec{z}_t)  \Big] \Bigg].
    \vspace{-0.2em}
\end{equation}

In practice, continuous diffusion models are trained to estimate the noise rather than directly predict the denoised data; this reparameterisation allows the loss to be easily minimised at time steps close to $T$. Unfortunately, no relevant reparameterisation currently exists for discrete distributions \cite{hoogeboom2021argmax}. Rather than directly maximising the ELBO, we reweight the ELBO to mimic the reparameterisation,
\vspace{-0.2em}
\begin{equation}\label{eqn:new-loss}
    \hspace*{-1em} \E_{q(\vec{z}_0)}\hspace*{-0.3em}\Bigg[ \sum_{t=1}^T\!\frac{T\!-\!t\!+\!1}{T}\!\E_{q(\vec{z}_t|\vec{z}_0)}\!\Big[ \sum_{[\vec{z}_t]_i=m}\hspace*{-0.7em}\log p_\theta([\vec{z}_0]_i|\vec{z}_t)  \Big]\!\Bigg]\!,\hspace*{-0.5em}
    \vspace{-0.2em}
\end{equation}
where components of the loss at time steps close to $T$ are weighted less than earlier steps. This is closely related to the loss obtained by assuming the posterior does not have access to $\vec{x}_t$, i.e. if the $t\!-\!1^\text{th}$ loss term is $D_{KL}( q(\vec{x}_{t-1}|\vec{x}_0) || p_\theta(\vec{x}_{t-1}|\vec{x}_t))$. Since we directly predict $\vec{z}_0$ and not $\vec{z}_t$ this assumption does not harm the training. Experimentally we find that this reweighting achieves lower validation ELBO than when directly maximising the ELBO.

Esser et al. \cite{esser2021taming} demonstrated that in the autoregressive case, Transformers \cite{vaswani2017attention} are better suited for modelling Vector-Quantized images than convolutional architectures due to the importance of long-distance relationships in this compressed form. As such, we utilise transformers to model the prior distribution, but without the architectural restrictions imposed by autoregressive approaches.

\subsection{Generating High-Resolution Images} \label{sec:extra-large-samples}
\noindent Using convolutions to build Vector-Quantized image models encourages latents to be highly spatially correlated with generated images. It is therefore possible to construct essentially arbitrarily sized images by generating latents with the required shape. We propose an approach that allows globally consistent images substantially larger than those in the training data to be generated. 

First, a large $a$ by $b$ array of mask tokens, $\bar{\vec{z}}_T = m^{a \times b}$, is initialised that corresponds to the size of image we wish to generate. In order to capture the maximum context when approximating $\bar{\vec{z}}_0$ we apply the denoising network to all subsets of $\bar{\vec{z}}_t$ with the same spatial size as the usual inputs of the network, aggregating estimates at each location. Specifically, using $c_j(\bar{\vec{z}}_t)$ to represent local subsets, we approximate the denoising distribution as a mixture,
\vspace{-0.1em}
\begin{equation}
    p([\bar{\vec{z}}_0]_i|\bar{\vec{z}}_t) \approx \frac{1}{Z} \sum_j p([\bar{\vec{z}}_0]_i|c_j(\bar{\vec{z}}_t)),
    \vspace{-0.2em}
\end{equation}
where the sum is over subsets $c_j$ that contain the $i^{th}$ latent. For extremely large images, this can require a very large number of function evaluations, however, the sum can be approximated by striding over latents with a step $>1$ or by randomly selecting positions. 

\begin{table}[t]
    \begin{tabular}{l>{\centering\arraybackslash}p{8mm}>{\centering\arraybackslash}p{8mm}>{\centering\arraybackslash}p{8mm}>{\centering\arraybackslash}p{8mm}}
        \toprule
        Model & P $\uparrow$ & R $\uparrow$ & D $\uparrow$ & C $\uparrow$ \\
        \midrule
        Churches & \\
        \quad DCT \cite{nash2021generating} & 0.60 & \textbf{0.48} & - & - \\ 
        \quad TT \cite{esser2021taming} & 0.67 & 0.29 & 1.08 & 0.60 \\ 
        \quad PGGAN \cite{karras2018progressive} & 0.61 & 0.38 & 0.83 & 0.63 \\
        \quad StyleGAN2 \cite{Karras2020analyzing} & 0.60 &	0.43 & 0.83 & 0.68 \\
        \quad \textbf{Ours ($t=1.0$)} & 0.70 & 0.42 & \textbf{1.12} & 0.73 \\
        \quad \textbf{Ours ($t=0.9$)} & \textbf{0.71} & 0.45 & 1.07 & \textbf{0.74} \\
        \midrule
        FFHQ & \\
        \quad VDVAE \cite{child2021very} & 0.59 & 0.20 & 0.80 & 0.50 \\
        \quad TT \cite{esser2021taming} & 0.64 & 0.29 & 0.89 & 0.59 \\
        \quad StyleGAN2 \cite{Karras2020analyzing} & 0.69 & 0.40 & 1.12 & 0.80 \\ 
        \quad \textbf{Ours ($t=1.0$)} & 0.69 & 0.48 & 1.06 & 0.77 \\
        \quad \textbf{Ours ($t=0.9$)} & \textbf{0.73} & \textbf{0.48} & \textbf{1.20} & \textbf{0.80}\\
        \midrule
        Bedroom & \\
        \quad DCT \cite{nash2021generating} & 0.44 & \textbf{0.56} & - & - \\ 
        \quad TT \cite{esser2021taming} & 0.61 & 0.33 & 1.15 & 0.75 \\
        \quad PGGAN \cite{karras2018progressive} & 0.43 & 0.40 & 0.70 & 0.64 \\
        \quad StyleGAN \cite{karras2019style} & 0.55 & 0.48 & 0.96 & 0.80 \\
        \quad \textbf{Ours ($t=1.0$)} & 0.64 & 0.38 & 1.27 & 0.81 \\ 
        \quad \textbf{Ours ($t=0.9$)} & \textbf{0.67} & 0.38 & \textbf{1.51} & \textbf{0.83} \\ 
        \bottomrule
    \end{tabular}
    \caption{Precision, Recall, Density, and Coverage for approaches trained on FFHQ, LSUN Bedroom, and LSUN Churches.}
    \label{tab:prdc-metrics}
    
    \vspace{0.9em}

    \begin{tabular}{llccc}
        \toprule
        Method & Params & Bed & Church & FFHQ \\
        \midrule
        DDPM \cite{ho2020denoising}           & 114M & 6.36 & 7.89 & - \\ 
        DCT \cite{nash2021generating}         & 448M & 6.40 & 7.56 & - \\
        VDVAE \cite{child2021very}            & 115M & -    & -    & 28.5 \\ 
        TT \cite{esser2021taming,esser2021imagebart} & 600M & 6.35 & 7.81 & 9.6 \\ 
        ImageBART \cite{esser2021imagebart}   & 2104M & 5.51 & 7.32 & 9.57 \\ 
        PGGAN \cite{karras2018progressive}    & 47M & 8.34 & 6.42 & - \\ 
        StyleGAN2 \cite{Karras2020analyzing}  & 60M & 2.35 & 3.86 & 3.8 \\ 
        \textbf{Ours ($t=1.0$)}                 & 145M & 5.07 & 5.58 & 7.12 \\ 
        \textbf{Ours ($t=0.9$)}                 & 145M & 3.64 & 4.07 & 6.11 \\ 
        \bottomrule
      \end{tabular}
    \caption{FID for various approaches on FFHQ, LSUN Bedroom, and LSUN Churches. Lower FID signifies higher quality samples.}
    \label{tab:dataset-fids}
    \vspace{-0.5em}
\end{table}

\begin{figure}[t]
    \centering
    \includegraphics[width=\linewidth]{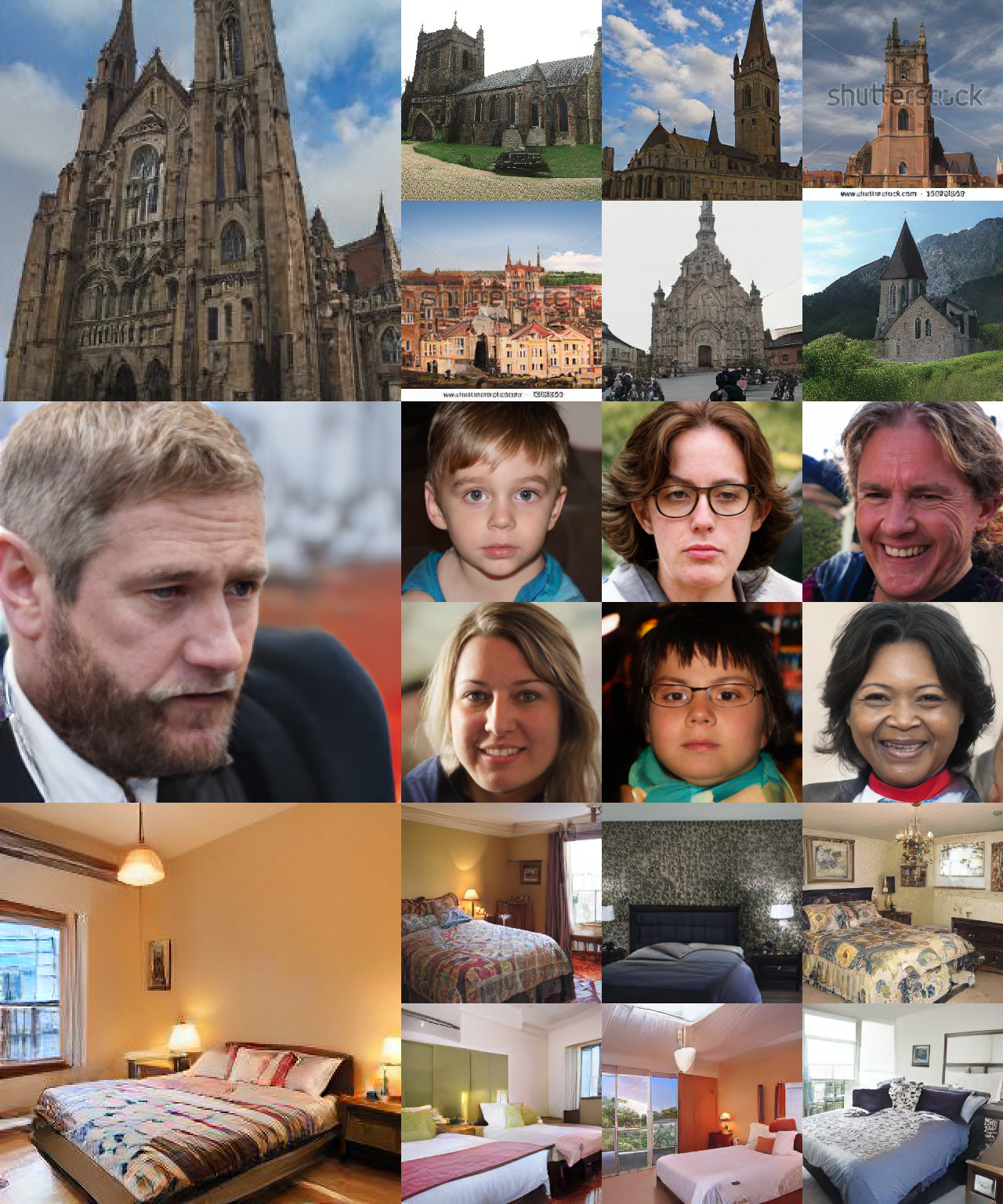}
    \caption{Samples from our models trained on 256x256 datasets: LSUN Churches, FFHQ, and LSUN Bedroom.}
    \label{fig:samples}
    \vspace{-0.5em}
\end{figure}

\subsection{Improving Code Representations}
\noindent There are various options to obtain high-quality image representations including using large numbers of latents and codes \cite{ramesh2021zero} or building a hierarchy of latent variables \cite{Razavi2019GeneratingDiverseHighFidelity}. 
We use the adversarial framework proposed by Esser et al. \cite{esser2021taming} to achieve higher compression rates with high-quality codes using only a single GPU, without tying our approach to the characteristics typically associated with generative adversarial models.
Additionally, we apply differentiable augmentations $T$, such as translations and colour jitter, to all discriminator inputs; this has proven to be effective at improving sample quality across methods \cite{jun2020distribution, zhao2020differentiable}. The overall loss $\loss$ is a linear combination of $\loss_\text{VQ}$, the Vector-Quantized loss, and $\loss_\text{G}$ which uses a discriminator $D$ to assess realism based on an adaptive weight $\lambda$. On some datasets, $\lambda$ can grow to extremely large values hindering training. We find simply clamping $\lambda$ at a maximum value $\lambda_\text{max}=1$ an effective solution that stabilises training,
\vspace{-0.4em}
\begin{subequations}
\begin{equation}\label{eqn:overall-loss}
    \loss = \min_{E,G,\mathcal{C}} \max_D \E_{\vec{x} \sim p_d} \big[ \loss_\text{VQ} + \lambda \loss_\text{G} \big],
\end{equation}
\begin{equation}\label{eqn:aug-gan-loss}
    \loss_\text{G} = \log D(T(\vec{x})) + \log(1 - D(T(\hat{\vec{x}}))),
\end{equation}
\begin{equation}\label{eqn:adaptive-weight}
    \lambda = \min \left(\frac{\nabla_{G_L}[\loss_\text{rec}]}{\nabla_{G_L}[\loss_\text{G}] + \delta}, \lambda_\text{max} \right).
\end{equation}
\end{subequations}

\section{Evaluation}
\vspace{-0.1em}
\noindent We evaluate our approach on three high-resolution 256x256 datasets: LSUN Bedroom, LSUN Churches \cite{yu2015lsun}, and FFHQ  \cite {karras2019style}. \cref{sec:experiments-sample-quality} evaluates the quality of samples from our proposed model. \cref{sec:experiments-diffusion} demonstrates the representation abilities of absorbing diffusion models applied to the learned discrete latent spaces, including how sampling can be sped up, improvements over equivalent autoregressive models, and the effect of our reweighted ELBO. Finally, \cref{sec:experiments-vqgan} evaluates our Vector-Quantized image model. 

\begin{figure*}[t]
    \centering
    \includegraphics[width=\linewidth]{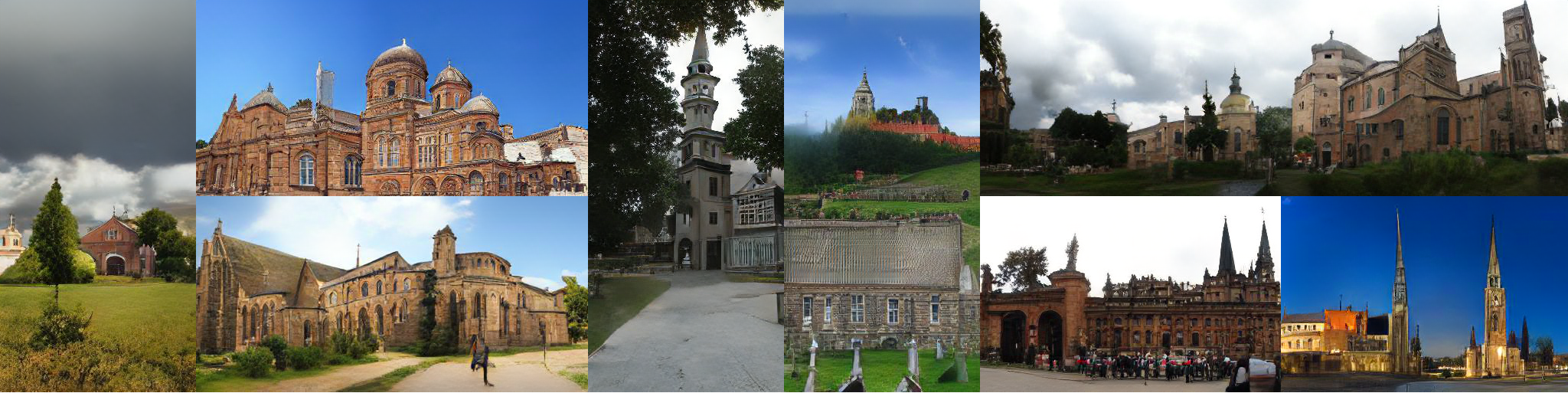}
    \caption{Our method allows unconditional images larger than those seen during training to be generated by applying the denoising network to all subsets of the image, aggregating probabilities to encourage global continuity.}
    \label{fig:big-samples}
\end{figure*}

In all experiments, our absorbing diffusion model parameterised with an 80M parameter Transformer Encoder \cite{vaswani2017attention} is applied to $16 \times 16$ latents discretised to a codebook with 1024 entries and optimised using the Adam optimiser \cite{kingma2014adam}. While, as noted by Esser et al. \cite{esser2021taming}, a GPT2-medium \cite{radford2019language} architecture (307M parameters) fits onto a GPU with 12GB of VRAM, in practice this requires the use of small batch sizes and learning rates making training in reasonable times impractical. More details can be found in \cref{apx:implementation-details}. Source code for the models used and experiments performed in this paper is available \href{https://github.com/samb-t/unleashing-transformers}{here}.

\subsection{Sample Quality} \label{sec:experiments-sample-quality}
\noindent In this section we evaluate samples from our model quantitatively and qualitatively. In comparison to other multi-step methods, our approach allows sampling in the fewest steps. Samples from our model can be found in \cref{fig:samples} which are high quality and diverse. More samples can be found in \cref{apx:more-samples}. 

\vspace{-1em}
\paragraph{PRDC} In \cref{tab:prdc-metrics} we evaluate our approach against a variety of other models in terms of Precision, Recall, Density, and Coverage (PRDC) \cite{sajjadi2018assessing, kynkaanniemi2019improved, naeem2020reliable}, metrics that quantify the overlap between the data and sample distributions. Due to limited computing resources, we are unable to provide density and coverage scores for DCT \cite{nash2021generating} and PRDC scores for StyleGAN2 on LSUN Bedroom since training on a standard GPU would take more than 30 days per experiment, significantly more than the 10 days required to train our models. On the LSUN datasets our approach achieves the highest Precision, Density, and Coverage; indicating that the data and sample manifolds have the most overlap. On FFHQ our approach achieves the highest Precision and Recall. In general, when generative models are sampled with lower temperatures to achieve lower FID, this leads to trade-off between precision and recall \cite{Karras2020analyzing, Razavi2019GeneratingDiverseHighFidelity}; since we also calculate FID with a lower temperature, we evaluate the effect of this on PRDC. In all but one case sampling with temperature leads to improved scores, indicating that our approach fits more accurately.

\vspace{-1em}
\paragraph{FID} In \cref{tab:dataset-fids} we calculate the Fr\'echet Inception Distance (FID) of samples from our models using torch-fidelity \cite{obukhov2020torchfidelity}. Despite using a fraction of the number of parameters compared to other Vector-Quantized image models, our approach achieves substantially lower FID scores. 

\vspace{-1em}
\paragraph{Higher Resolution} \cref{fig:big-samples} contains samples of various spatial sizes using the approach described in \cref{sec:extra-large-samples}. Here, an absorbing diffusion model is trained on $16\!\times\!16$ latents then samples are generated at larger sizes (up to $768\!\times\!256$) using a temperature value of 0.8.  Even at the larger scales we observe high-quality, diverse, and consistent imagery.

\vspace{-0.5em}
\subsubsection{Limitations of FID Metric}
While FID has been found to correlate well with image quality, it unrealistically approximates the data distribution as Gaussian in embedding space and is insensitive to the global structure of the data distribution \cite{tsitsulin2020shape}. For likelihood models, calculating NLL on a test set is possible instead but likelihood has been shown to not correlate well with quality; fine tuning our approach to model pixels as Gaussians gives 2.72BPD on 5-bit FFHQ. Alternative approaches that address these issues have been developed \cite{borji2021pros} such as PPL \cite{karras2019style}, which assesses sample consistency through latent interpolations; IMD  \cite{tsitsulin2020shape}, which uses all moments to compare data manifolds making it sensitive to global structure; and MTD \cite{barannikov2021manifold}, which compares manifolds in image space.

In this work, we compare approaches using Precision and Recall \cite{sajjadi2018assessing} approaches which, unlike FID, evaluate sample quality and diversity separately and have been used in similar recent work assessing high-resolution image generation \cite{Karras2020analyzing, nash2021generating, hudson2021ganformer, Razavi2019GeneratingDiverseHighFidelity}. Precision is the expected likelihood of fake samples lying on the data manifold and recall vice versa. These metrics are computed by approximating the data and sample manifolds as hyper-spheres around data and sample points respectively \cite{kynkaanniemi2019improved}. Density and Coverage are modifications to Precision and Recall respectively that address manifold overestimation \cite{naeem2020reliable}.

\subsection{Absorbing Diffusion} \label{sec:experiments-diffusion}
\noindent In this section we analyse the usage of absorbing diffusion for high-resolution image generation, determining how many sampling steps are required to obtain high-quality samples and ablating the components of our approach.

\begin{table}[b]
    \centering
    \begin{tabular}{cccccc}
        \toprule
        Steps    & 50   & 100   & 150    & 200   & 256  \\
        \midrule
        Churches & 6.86 & 6.09  & 5.81   & 5.68  & 5.58 \\
        Bedroom  & 6.85 & 5.83  & 5.53   & 5.32  & 5.42 \\
        FFHQ     & 9.60 & 7.90  & 7.53   & 7.52  & 7.12 \\
        \bottomrule
    \end{tabular}
    \caption{FID for different number of sampling steps on LSUN Churches, Bedroom and FFHQ. Diffusion steps are evenly spaced.}
    \label{tab:fast-sampling}
\end{table}

\vspace{-0.5em}
\subsubsection{Sampling Speed}
\vspace{-0.1em}
Our approach applies a diffusion process to a highly compressed image representation, meaning it is already $18\times$ faster to sample from than DDPM (ours: 3.8s, DDPM: 70s per image on a NVIDIA RTX 2080 Ti). However, since the absorbing diffusion model is trained to approximate $p(\vec{z}_0|\vec{z}_t)$ it is possible to speed the sampling process up further by skipping arbitrary numbers of time steps, unmasking multiple latents at once. In \cref{tab:fast-sampling} we explore how sample quality is affected using a simple step skipping scheme: evenly skipping a constant number of steps so that the total number of steps meets some fixed computational budget. As expected, FID increases with fewer sampling steps. However, the increase in FID is minor relative to the improvement in sampling speed: our approach achieves similar FID to the equivalent autoregressive model using half the number of steps. With 50 sampling steps, our approach is $88\times$ faster than DDPMs. Using a more sophisticated step selection scheme such as dynamic programming \cite{watson2021learning}, FID could potentially be reduced further.

\vspace{-0.5em}
\subsubsection{Autoregressive vs Absorbing DDPM}
\vspace{-0.1em}
\cref{tab:fid-ar-vs-ddpm} compares the representation ability of our absorbing diffusion model with an autoregressive model, both utilising  exactly the same Transformer architecture, but with the  Transformer unconstrained in the diffusion case. On both datasets tested, the diffusion achieves lower FID and NLL than the autoregressive model despite being trained on a harder task with the same number of parameters; the additional regularisation prevents overfitting which is prevalent with autoregressive models \cite{esser2021imagebart, jun2020distribution}. Since the number of sampling steps is the same as the number of data dimensions, samples from the diffusion are effectively autoregressive over random orderings, indicating that the learned distributions better approximate the data distribution.

\begin{table}[t]
    \centering
    \begin{tabular}{lcccc}
        \toprule
        \multirow{2}{*}{Method} & \multicolumn{2}{c}{Churches} & \multicolumn{2}{c}{FFHQ} \\
                       & FID $\downarrow$ & NLL $\downarrow$ & FID $\downarrow$ & NLL $\downarrow$ \\
        \midrule
        Autoregressive & 5.93 & 6.24 & 8.15  & 6.18 \\ 
        Absorbing DDPM & \textbf{5.58} & \textbf{6.01} & \textbf{7.12} & \textbf{5.96} \\
        \bottomrule
    \end{tabular}
    \caption{FID and validation NLL (in BPD) for different methods to approximate discrete latents using the same Transformer architecture on LSUN Churches and FFHQ.} 
    \label{tab:fid-ar-vs-ddpm}
\end{table}

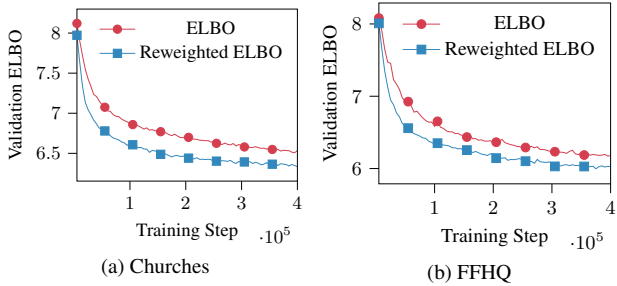
\begin{figure}[t]
    \centering
    \begin{subfigure}{0.49\linewidth}
        \centering
        \begin{adjustbox}{width=\linewidth}
            \begin{tikzpicture}
            \tikzstyle{every node}=[font=\small]
            \begin{axis}[
                width=1.3\linewidth,height=1.1\linewidth,
                xlabel={Training Step},
                ylabel={Validation ELBO},
                xtick pos=left,
                ytick pos=left,
                enlarge x limits=false,
                every x tick/.style={color=black, thin},
                every y tick/.style={color=black, thin},
                tick align=outside,
                xlabel near ticks,
                ylabel near ticks,
                axis on top,
                legend style={draw=none},
            ]
            \addplot+[spectral2, mark options={fill=spectral2}, mark repeat=10] table [x=step, y=normed, col sep=comma] {plots/churches_loss_comparison.csv};\addlegendentry{ELBO}
            \addplot+[spectral10, mark options={fill=spectral10}, mark repeat=10] table [x=step, y=new, col sep=comma] {plots/churches_loss_comparison.csv};\addlegendentry{Reweighted ELBO}
            
            \end{axis}
            \end{tikzpicture}
        \end{adjustbox}
        \caption{Churches}
    \end{subfigure}
    \begin{subfigure}{0.49\linewidth}
        \centering
        \begin{adjustbox}{width=\linewidth}
            \begin{tikzpicture}
            \tikzstyle{every node}=[font=\small]
            \begin{axis}[
                width=1.3\linewidth,height=1.1\linewidth,
                xlabel={Training Step},
                ylabel={Validation ELBO},
                xtick pos=left,
                ytick pos=left,
                enlarge x limits=false,
                every x tick/.style={color=black, thin},
                every y tick/.style={color=black, thin},
                tick align=outside,
                xlabel near ticks,
                ylabel near ticks,
                axis on top,
                legend style={draw=none},
            ]
            \addplot+[spectral2, mark options={fill=spectral2}, mark repeat=10] table [x=step, y=elbo, col sep=comma] {plots/ffhq_loss_comparison.csv};\addlegendentry{ELBO}
            \addplot+[spectral10, mark options={fill=spectral10}, mark repeat=10] table [x=step, y=new, col sep=comma] {plots/ffhq_loss_comparison.csv};\addlegendentry{Reweighted ELBO}
            \end{axis}
            \end{tikzpicture}
        \end{adjustbox}
        \caption{FFHQ}
    \end{subfigure}
    \caption{Comparison of proposed losses on (a) LSUN Churches and (b) FFHQ. Models trained with our reweighted ELBO achieve lower validation ELBO than models trained directly on the ELBO.}
    \label{fig:elbo-vs-rewighted}
\end{figure}

\vspace{-0.5em}
\subsubsection{Reweighted ELBO}
\vspace{-0.1em}
In \cref{sec:method-global} we proposed using a reweighted ELBO when training the diffusion model that focuses gradients on the central training steps, balancing feasability with ease of learning. We evaluate this in \cref{fig:elbo-vs-rewighted} by comparing validation ELBO during training for models trained directly on ELBO and our re-weighting. The re-weighted ELBO converges to a lower validation ELBO sooner, demonstrating that our reweighting is valid and simplifies optimisation. 

\begin{figure}[t]
    \centering
    \begin{tikzpicture}
    \node[anchor=south west,inner sep=0] (image) at (0,0) {\includegraphics[width=\linewidth]{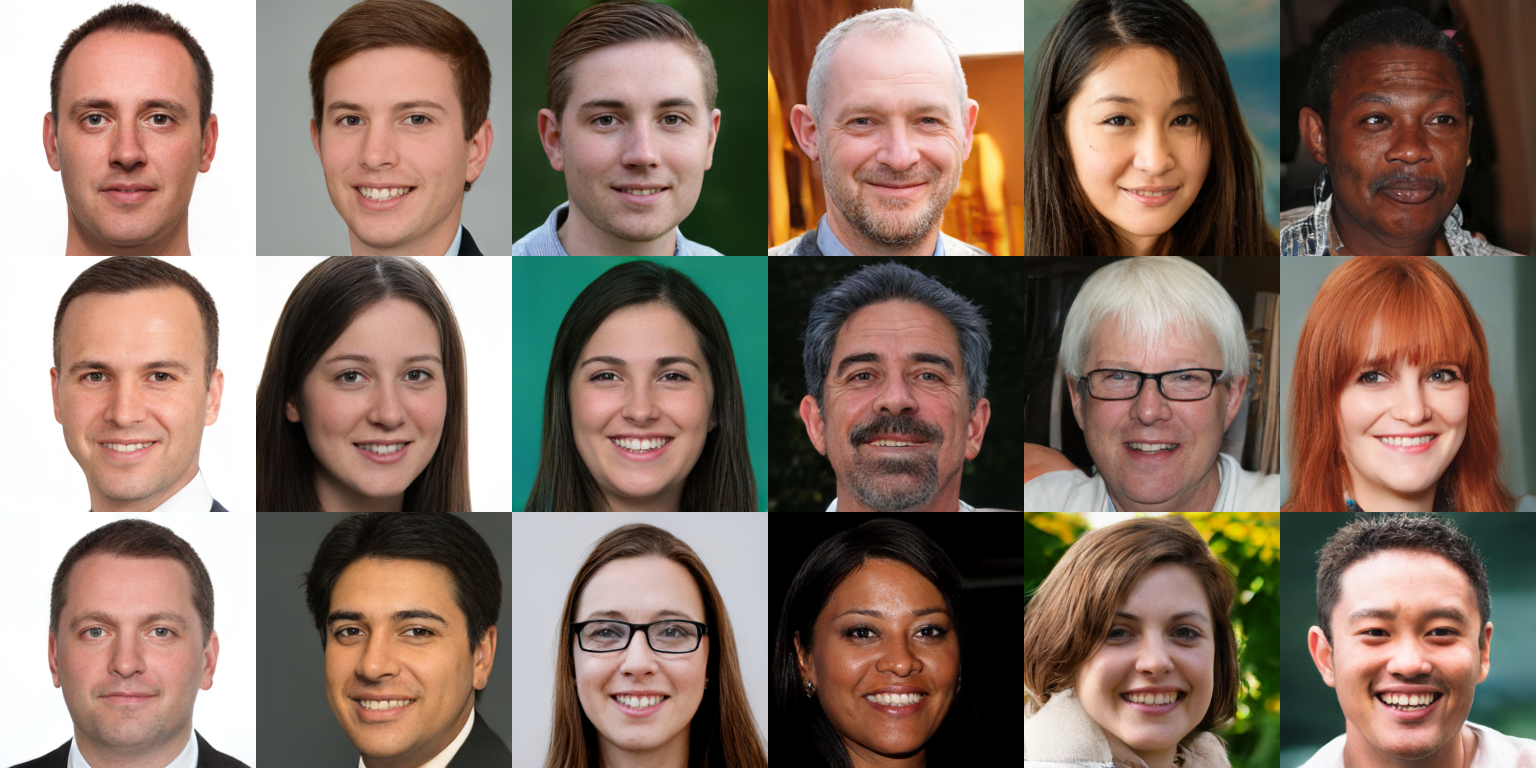}};
    \draw[-latex] (0,-0.15) -- (8.34,-0.15);
    \node[align=center] at (4.17,-0.4) {\small Temperature};
    \end{tikzpicture}
    \vspace{-1.7em}
    \caption{Impact of sampling temperature on diversity. For small temperature changes it is less obvious how bias has changed.}
    \label{fig:temp-samples}
\end{figure}

\subsection{Sample Diversity}
\vspace{-0.07em}
\noindent To improve sample quality, many generative models are sampled using a reduced temperature or by truncating Normal distributions. This is problematic, as these sampling methods will amplify any biases in the dataset. We visualise the impact of temperature on sampling from a model trained on FFHQ in \cref{fig:temp-samples}. For very low temperatures the bias the obvious: samples are mostly front-facing white men with brown hair on solid white/black backgrounds. Exactly how the bias has changed for more subtle temperature changes is less clear, which is problematic. Practitioners should be aware of this effect and it emphasises the importance of dataset balancing.

\begin{table}[b]
    \centering
    \begin{tabular}{lcc}
        \toprule
        Modifications & Churches & FFHQ  \\
        \midrule
        Default & 5.25 & 3.37 \\
        $\lambda_\text{max}=1$ & 8.67 & 4.72 \\
        DiffAug & 5.16 & 6.57 \\
        Both & \textbf{2.70} & \textbf{3.12} \\
        \bottomrule
    \end{tabular}
    \caption{Effect of DiffAug and adaptive weight limiting on reconstruction FID. Results were calculated on Vector-Quantized image models trained on LSUN Churches and FFHQ for 500k steps.}
    \label{tab:vqgan-modifications}
\end{table}

\subsection{Reconstruction Quality} \label{sec:experiments-vqgan}
\vspace{-0.07em}
\noindent In \cref{tab:vqgan-modifications} we evaluate the effect of differentiable augmentations (DiffAug) \cite{zhao2020differentiable} and adaptive weight limiting on Vector-Quantized image modelling. While applying each technique individually can lead to worse FID due to imbalance, when both techniques are applied, we found that FID improved across all datasets tested.

\subsection{Image Editing} 
\begin{figure}[t]
    \centering
    \begin{picture}(250,96)
    \put(0,0){\includegraphics[width=\linewidth]{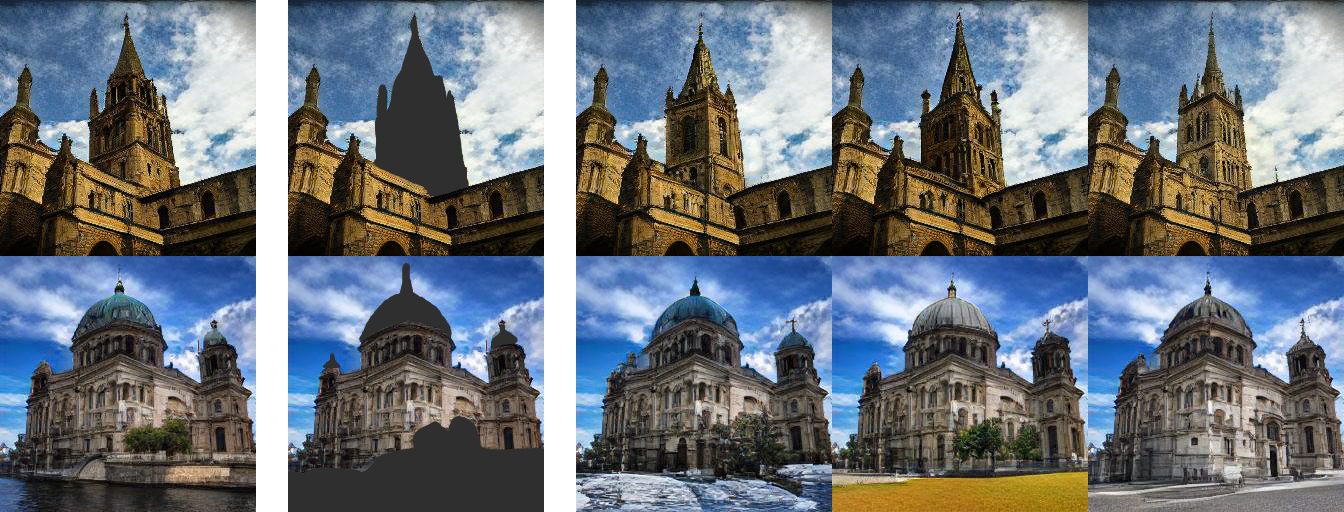}}
    \put(8,94){\small Original}
    \put(59,94){\small Masked}
    \put(155,94){\small Outputs}
    \end{picture}
    \caption{Our approach allows local image editing by targeting regions to be changed (highlighted in grey). Here alternative towers are generated and water is replaced with different foregrounds.}
    \label{fig:inpainting}
\end{figure}
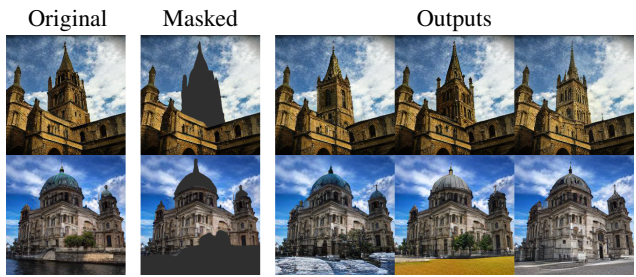

\noindent An additional advantage of using a bidirectional diffusion model to model the latent space is that image inpainting is possible. Since autoregressive models are conditioned only on the upper left region of the image, they are unable to edit internal masked image regions in a consistent manner. Diffusion models, on the other hand, allow masked regions to be placed at arbitrary locations. After a region has been highlighted, we mask corresponding latents, identify the starting time step by counting the number of masked latents, then continue the denoising process from that point. Examples of this process can be found in \cref{fig:inpainting}.

\subsection{Limitations}
\noindent In our experiments we only tested our approach on $256\!\times\!256$ datasets; directly scaling to higher resolutions would require more GPU resources. However, future work using more efficient Transformer architectures \cite{jaegle2021perceiver} may alleviate this. Our method outperforms all approaches tested on FID except StyleGAN2 \cite{Karras2020analyzing}; we find that the primary bottleneck is the Vector-Quantized image model, therefore more research is necessary to improve these discrete representations. Whilst our approach is trained for significantly less time than other approaches such as StyleGAN2, the stochastic training procedure means that more training steps are required compared to autoregressive approaches. Although when generating extra-large images the large context window made possible by the diffusion model encourages consistency, a reduced temperature is also required, reducing diversity.

\section{Discussion}
\noindent While other classes of discrete generative model exist, they are less suitable for Vector-Quantized image modelling than discrete diffusion models: VAEs introduce prior assumptions about the latent space that can be limiting, in particular, continuous spaces may not be appropriate when modelling discrete data \cite{bowman2016generating}; GAN training requires sampling from the generator meaning that gradients must be backpropagated through a discretistion procedure \cite{nie2019relgan}; Discrete normalising flows require the use of invertible functions which significantly restrict the function space \cite{hoogeboom2019integer, berg2020idf++}.

Concurrently developed with this work was ImageBART \cite{esser2021imagebart} which also uses a diffusion model to learn the prior of a Vector-Quantized image model. However, our approaches substantially differ: ImageBART uses a multinomial diffusion process with separate autoregressive Transformers trained to approximate each diffusion step, leading to slower inference and substantially more parameters; our approach optimises all diffusion steps simultaneously with a single, non-autoregressive Transformer. 

There are numerous potential avenues to explore to further improve sample quality. For instance, when learning discrete image representations, implicit networks (which are invariant to translation and rotation) \cite{karras2021alias} or other more powerful generative models could be used. Alternatively, different discrete diffusion methods could be used that impose relationships between codes based on their continuous embeddings. Finally, by conditioning on both text and discrete image representations, absorbing diffusion models could allow text-to-image generation and image captioning to be accomplished using a single model with faster run-time than independent approaches \cite{ramesh2021zero, radford2021learning}.

\subsection{Social Impact}
\noindent While deep generative models have various positive applications such as text-to-speech, drug design, and generating examples of rare medical conditions, there can be negative consequences associated with their development:
\begin{itemize}[itemsep=2pt,parsep=0pt,topsep=2pt]
    \item As with all generative models, our approach can be used for malicious applications such as deep-fakes.
    \item Samples reflect the biases present in the training data which can lead to unintended consequences.
    \item Training generative models such as ours consumes significant quantities of energy affecting the environment. The fast sampling permitted by our approach, however, reduces this at test-time compared to similar methods.
\end{itemize}
\section{Conclusion}
\noindent In this work we proposed a discrete diffusion probabilistic model prior capable of predicting Vector-Quantized image representations in parallel, overcoming the high sampling times, unidirectional nature and overfitting challenges associated with autoregressive priors. Our approach makes no assumptions about the inherent ordering of latents by utilising an unconstrained Transformer architecture. Experimental results demonstrate the ability of our approach to generate diverse, high-quality images, optionally at resolutions exceeding the training samples. Additional work is needed to reduce training times and to efficiently scale our approach to even higher resolutions.

{\small
\bibliographystyle{ieee_fullname}
\bibliography{main_bib}
}

\newpage \pagebreak \cleardoublepage
\newpage
\appendix
\begin{center}
    \large{\textbf{Supplementary Material}}
\end{center}

\noindent The supplementary material for this work is divided into the following sections: Section \ref{apx:implementation-details} describes the architectures and hyperparameters for the experiments presented in the main paper; Section \ref{apx:reweighted-elbo} illustrates the connection between our proposed ELBO reweighting and the true ELBO; Section \ref{apx:nearest-neighbours} gives nearest neighbour examples to demonstrate generalisation; and finally, Section \ref{apx:more-samples} contains a large number of uncurated samples.

\section{Implementation Details} \label{apx:implementation-details}

\noindent We perform all experiments on a single NVIDIA RTX 2080 Ti with 11GB of VRAM using automatic mixed precision when possible. As mentioned in the main paper, we use the same VQGAN architecture as used by Esser et al. \cite{esser2021taming} which for $256 \times 256$ images downsamples to features of size $16 \times 16 \times 256$, and quantizes using a codebook with 1024 entries. Attention layers are applied within both the encoder and decoder on the lowest resolutions to aggregate context across the entire image. Models are optimised using the Adam optimiser \cite{kingma2014adam} using a batch size of 4 and learning rate of $1.8\times10^{-5}$. For the differentiable augmentations we randomly change the brightness, saturation, and contrast, as well as randomly translate images. The datasets we use are both publically accessible, with FFHQ available under the Creative Commons BY 4.0 licence. LSUN models are trained for 2.2M steps while the FFHQ model is trained for 1.4M steps.

For the absorbing diffusion model we use a scaled down 80M parameter version of GPT-2 \cite{radford2019language} consisting of 24 layers, where each attention layer has 8 heads, each 64D. The same architecture is used for experiments with the autoregressive model. Autoregressive models' training are stopped based on the best validation loss. We also stop training the absorbing diffusion models based on validation ELBO, however, on the LSUN datasets we found that it always improved or remained consistent throughout training so each model was trained for 2M steps.

\paragraph{Codebook Collapse} One issue with vector quantized methods is codebook collapse, where some codes fall out of use which limits the potential expressivity of the model. We found this to occur across all datasets with often a fraction of the codes in use. We experimented with different quantization schemes such as gumbel softmax, different initalisation schemes such as k-means, and `code recycling', where codes out of use are reset to an in use code. In all of these cases, we found the reconstruction quality to be comparable or worse so stuck with the argmax quantisation scheme used by Esser et al. \cite{esser2021taming}.

\paragraph{Precision, Recall, Density, and Coverage} To compute these measures we use the official code releases and pretrained weights in all cases except Taming Transformers on the LSUN datasets where weights were not available; in this case we reproduced results as close as possible with the hardware available, training the VQGANs and autoregressive models with the same hyperparameters used for the rest of our experiments. Following Nash et al. \cite{nash2021generating} we use the standard 2048D InceptionV3 features, which are also used to compute FID. The measures are computed using the code provided by Naeem et al. \cite{naeem2020reliable}. 

\section{Reweighted ELBO} \label{apx:reweighted-elbo}
\noindent In \Cref{sec:method-global} we propose re-weighting the ELBO of the absorbing diffusion model so that the individual loss at each time step is multiplied by $\frac{T-t+1}{T}$ rather than $1/t$. In this section we justify the correctness of this re-weighting by showing it is equivalent to minimising the difference to a forward process that does not have access to $\vec{x}_t$. As such, the loss takes into account the difficulty of denoising steps and re-weights them down accordingly. In this case, the loss at time step $t$ can be written as
\begin{equation}\label{eqn:new-loss-kld}
    \begin{split}
        \loss_t &= D_\text{KL}(q(\vec{x}_{t-1}|\vec{x}_0) || p(\vec{x}_{t-1}|\vec{x}_t)) \\
        &= \sum_i \sum_j q([\vec{x}_{t-1}]_{i,j} | \vec{x}_0) \log \frac{q([\vec{x}_{t-1}]_{i,j}|\vec{x}_0)}{p([\vec{x}_{t-1}]_{i,j}|\vec{x}_t)},
    \end{split}
\end{equation}
where the first summation sums over latent coordinates $i$, and the second summation sums over the probabilities of each code $j$. For the absorbing diffusion case where tokens in $\vec{x}_t$ are masked independently and uniformly with probability $\frac{t}{T}$, this posterior is defined as
\begin{equation}
    \begin{split}
        q([\vec{x}_{t-1}]_i = a| \vec{x}_0) & \\
        &\hspace{-6em} = \begin{cases}
            1 - \frac{t-1}{T}, & \text{if $a=[\vec{x}_0]_i$ and $[\vec{x}_t]_i=m$}.\\
            \frac{t-1}{T}, & \text{if $a=m$ and $[\vec{x}_t]_i=m$}. \\
            1, & \text{if $a=[\vec{x}_0]_i$ and $[\vec{x}_t]_i=[\vec{x}_0]_i$}.\\
            0, & \text{otherwise}.
        \end{cases}
    \end{split}
\end{equation}
The reverse process remains defined in the same way as the standard reverse process:
\begin{equation}
    \begin{split}
      p([\vec{x}_{t-1}]_i = a| \vec{x}_t) & \\ 
      &\hspace{-7em} = \begin{cases}
        \frac{1}{t}p_\theta([\vec{x}_0]_i|\vec{x}_t), & \text{if $a=[\vec{x}_0]_i$ and $[\vec{x}_t]_i=m$}.\\
        1 - \frac{1}{t}, & \text{if $a=m$ and $[\vec{x}_t]_i=m$}. \\
        1, & \text{if $a=[\vec{x}_0]_i$ and $[\vec{x}_t]_i=[\vec{x}_0]_i$}.\\
    \end{cases}\hspace{-1em}
  \end{split}
\end{equation}
Substituting these definitions into \Cref{eqn:new-loss-kld}, the loss can be simplified to \Cref{eqn:new-loss-almost}; by extracting the constants into a single term out of the sum, $C$, the loss can be further simplified to obtain \Cref{eqn:new-loss-constant}, which is equivalent to our proposed reweighted ELBO \Cref{eqn:new-loss-kld},
\begin{equation}\label{eqn:new-loss-almost}
    \begin{split}
        \loss_t &= \sum_i \bigg[ 1 \log \frac{1}{1} + \frac{t-1}{T}\log\frac{\frac{t-1}{T}}{1-\frac{1}{t}} \\ &\hspace{3.25em} + \left( 1 - \frac{t-1}{T} \log \frac{1-\frac{t-1}{T}}{\frac{1}{t}p_\theta([\vec{x}_0]_i|\vec{x}_t)} \right) \bigg],
    \end{split}
\end{equation}
\begin{equation}\label{eqn:new-loss-constant}
    = C - \sum_i \left[  \frac{T-t-1}{T}  \log p_\theta([\vec{x}_0]_i|\vec{x}_t)  \right].
\end{equation}

\section{Nearest Neighbours}\label{apx:nearest-neighbours}
\noindent When training generative models, being able to detect overfitting is key to ensure the data distribution is well modelled. Overfitting is not detected by popular metrics such as FID, making overfitting difficult to identify in approaches such as GANs. With our approach we are able to approximate the ELBO on a validation set making it simple to prevent overfitting. In this section we demonstrate that our approach is not overfit by providing nearest neighbour images from the training dataset to samples from our model, measured using LPIPS \cite{zhang2018unreasonable}.

\newpage
\section{Additional Samples} \label{apx:more-samples}
\begin{figure*}[h]
    \centering
    \includegraphics[width=\textwidth]{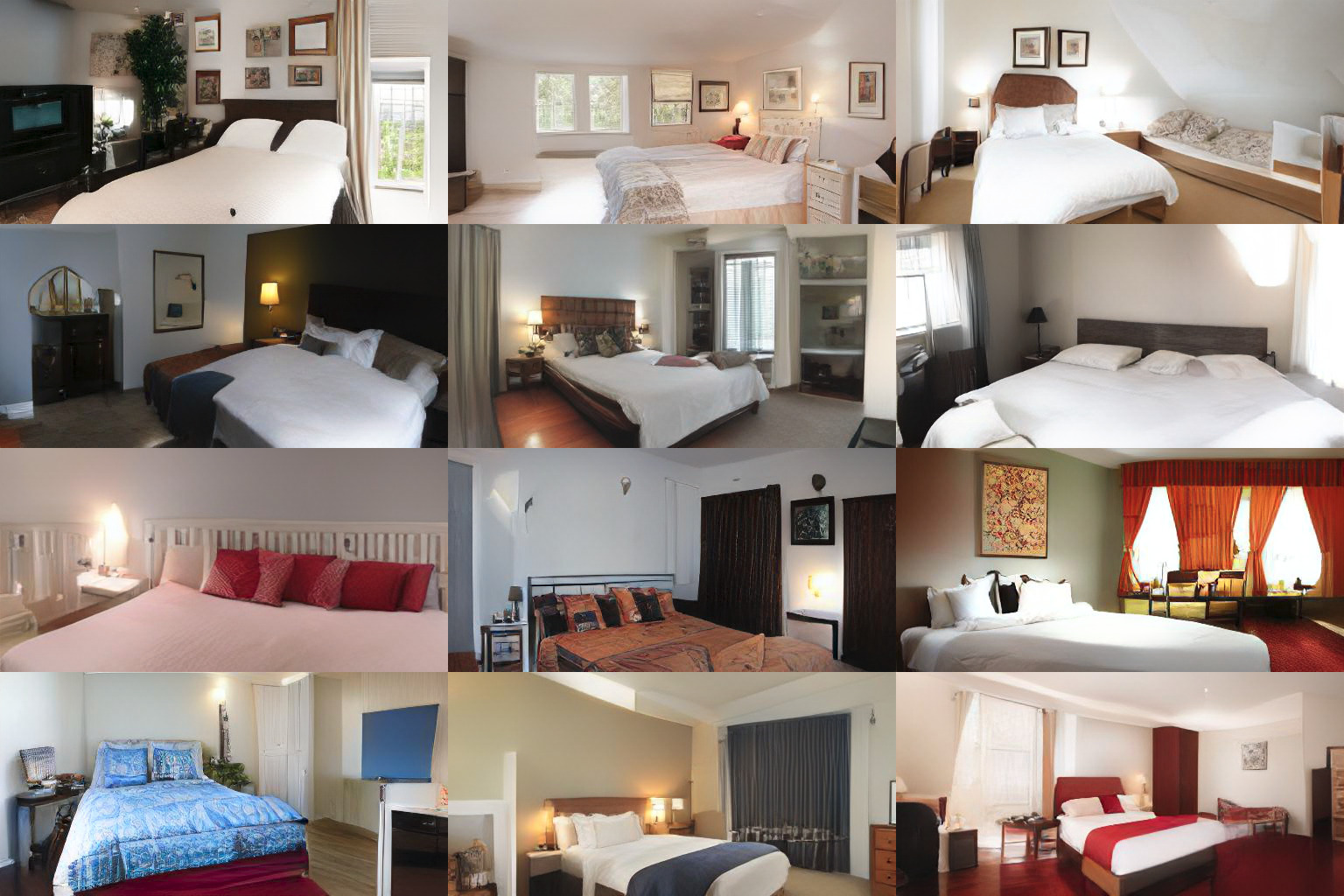}
    \caption{Unconditional samples from a model trained on LSUN Bedroom larger than images in the training dataset.}
    \label{fig:many-big-bedrooms}
\end{figure*}

\begin{figure*}[h]
    \centering
    \includegraphics[width=\textwidth]{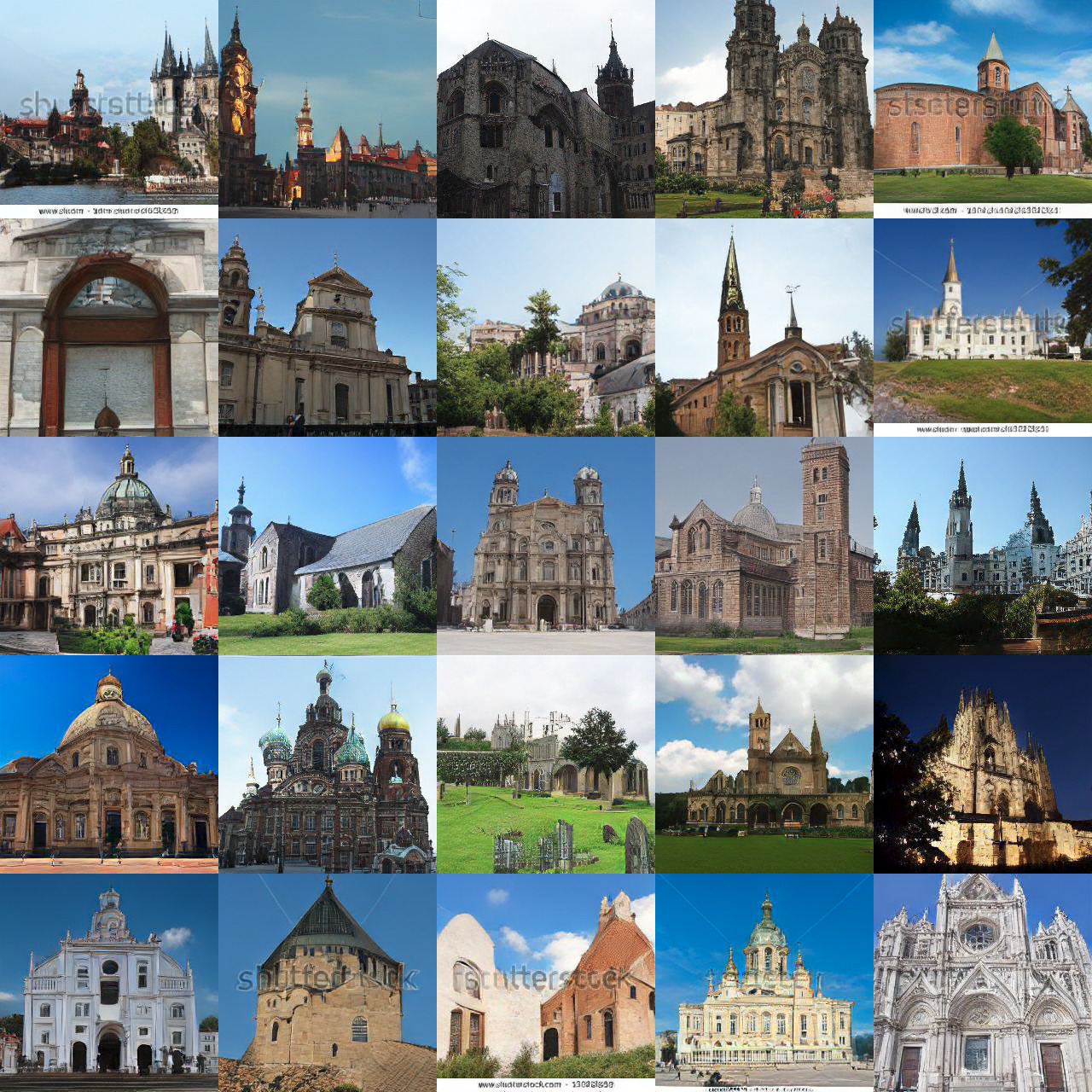}
    \caption{Non-cherry picked, $t=0.9$, 256x256 LSUN Churches samples.}
    \label{fig:many-church-samples}
\end{figure*}

\begin{figure*}[h]
    \centering
    \includegraphics[width=\textwidth]{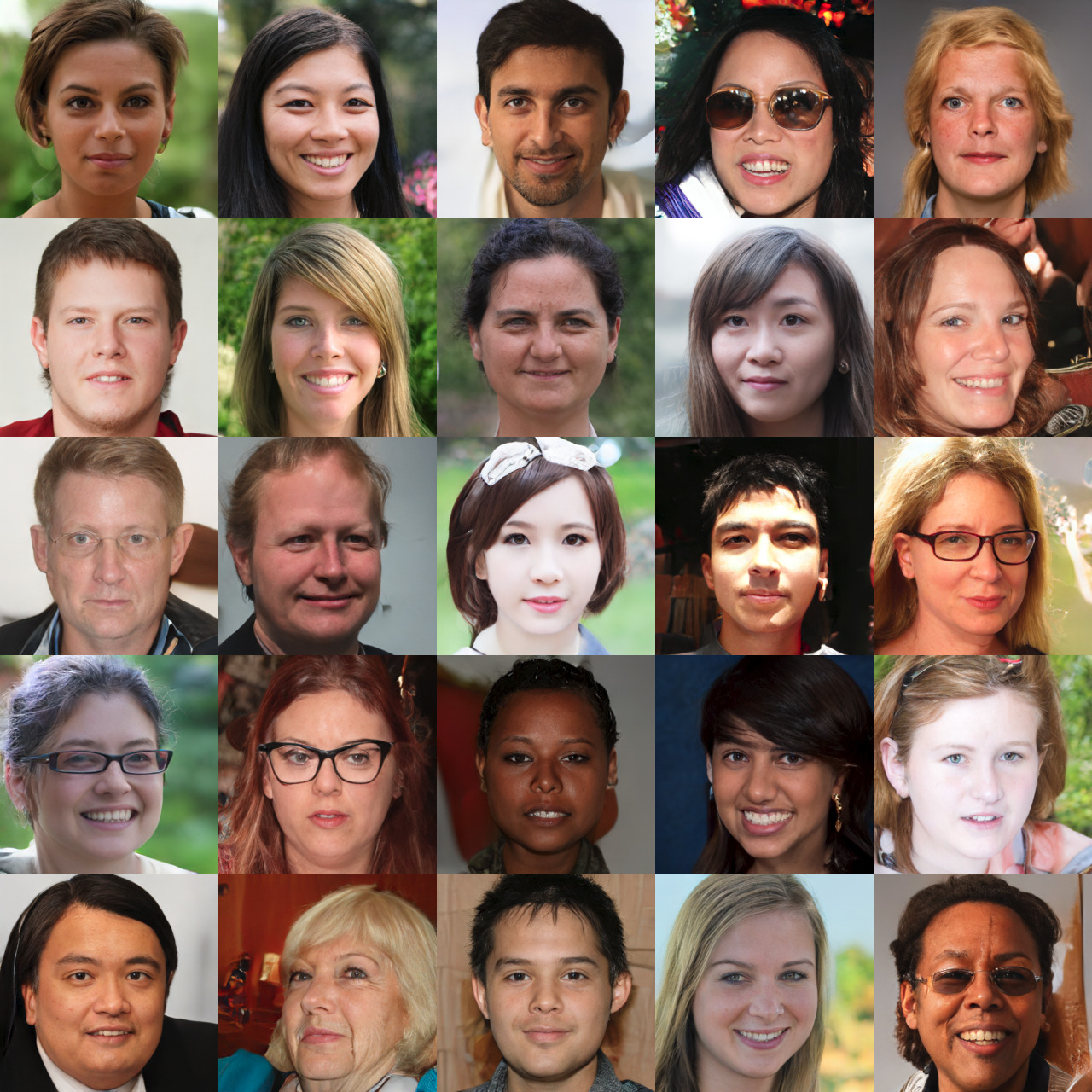}
    \caption{Non-cherry picked, $t=0.85$, 256x256 FFHQ samples.}
    \label{fig:many-ffhq-samples}
\end{figure*}

\begin{figure*}[h]
    \centering
    \includegraphics[width=\textwidth]{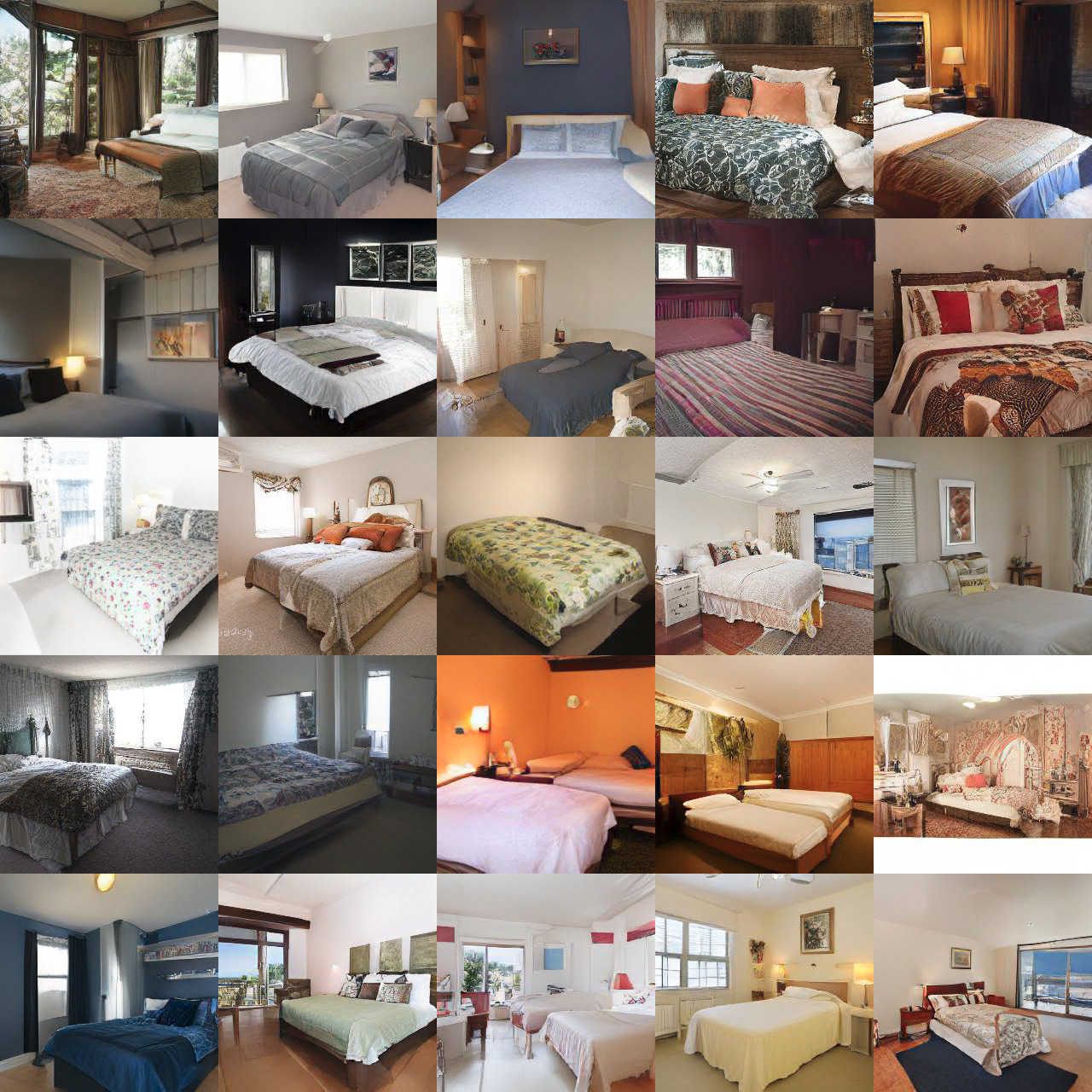}
    \caption{Non-cherry picked, $t=0.9$, 256x256 LSUN Bedroom samples.}
    \label{fig:many-bedroom-samples}
\end{figure*}

\noindent \Cref{fig:many-big-bedrooms} contains unconditional samples with resolutions larger than observed in the training data from a model trained on LSUN Bedroom. In \Cref{fig:many-ffhq-samples,fig:many-church-samples,fig:many-bedroom-samples} additional samples from our models are visualised on LSUN Churches, LSUN Bedroom, and FFHQ.

\begin{figure*}[b]
    \centering
    \includegraphics[width=\textwidth]{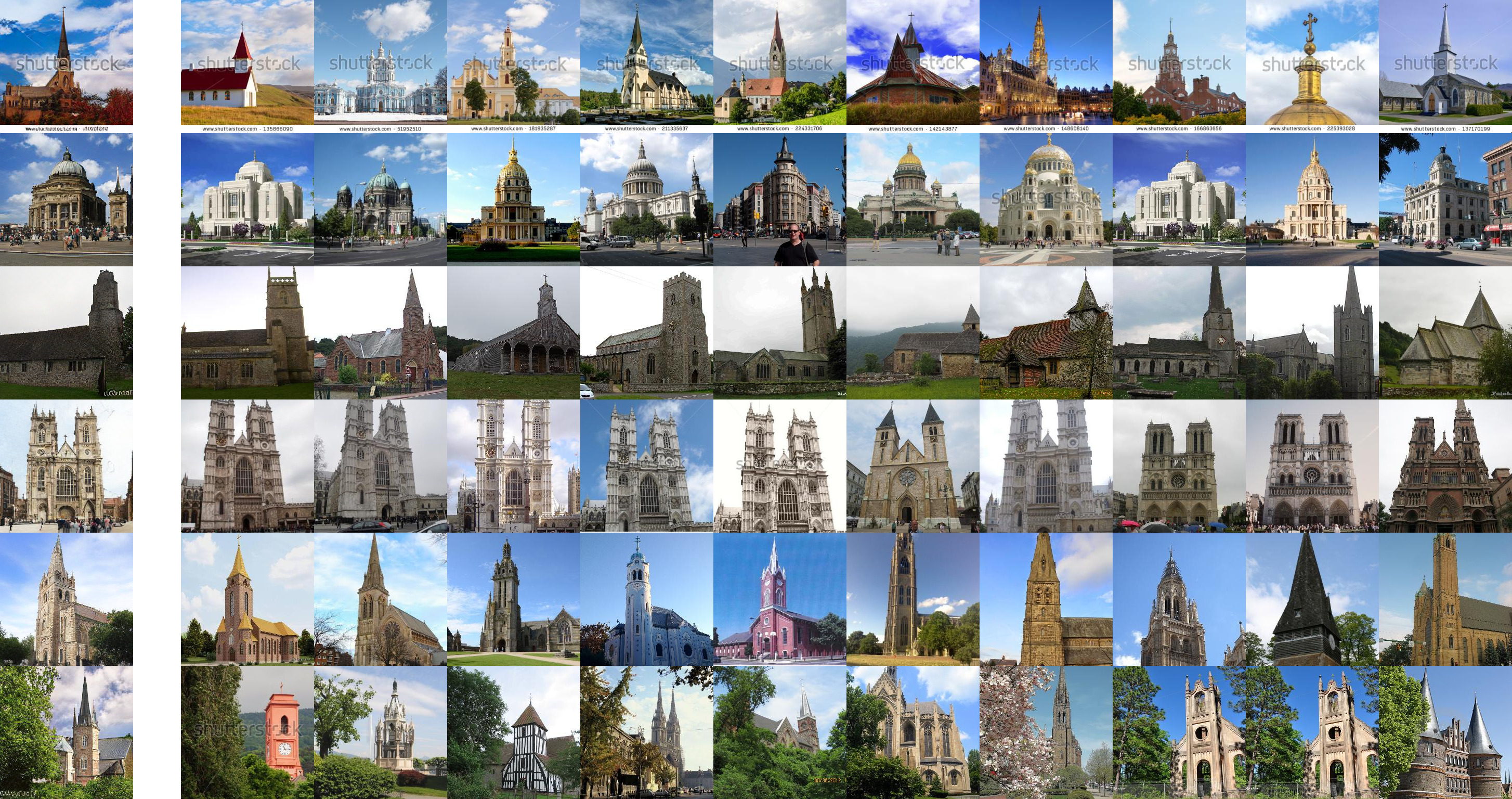}
    \caption{Nearest neighbours for a model trained on LSUN Churches based on LPIPS distance. The left column contains samples from our model and the right column contains the nearest neighbours in the training set (increasing in distance from left to right).}
\end{figure*}

\begin{figure*}[h]
    \centering
    \includegraphics[width=\textwidth]{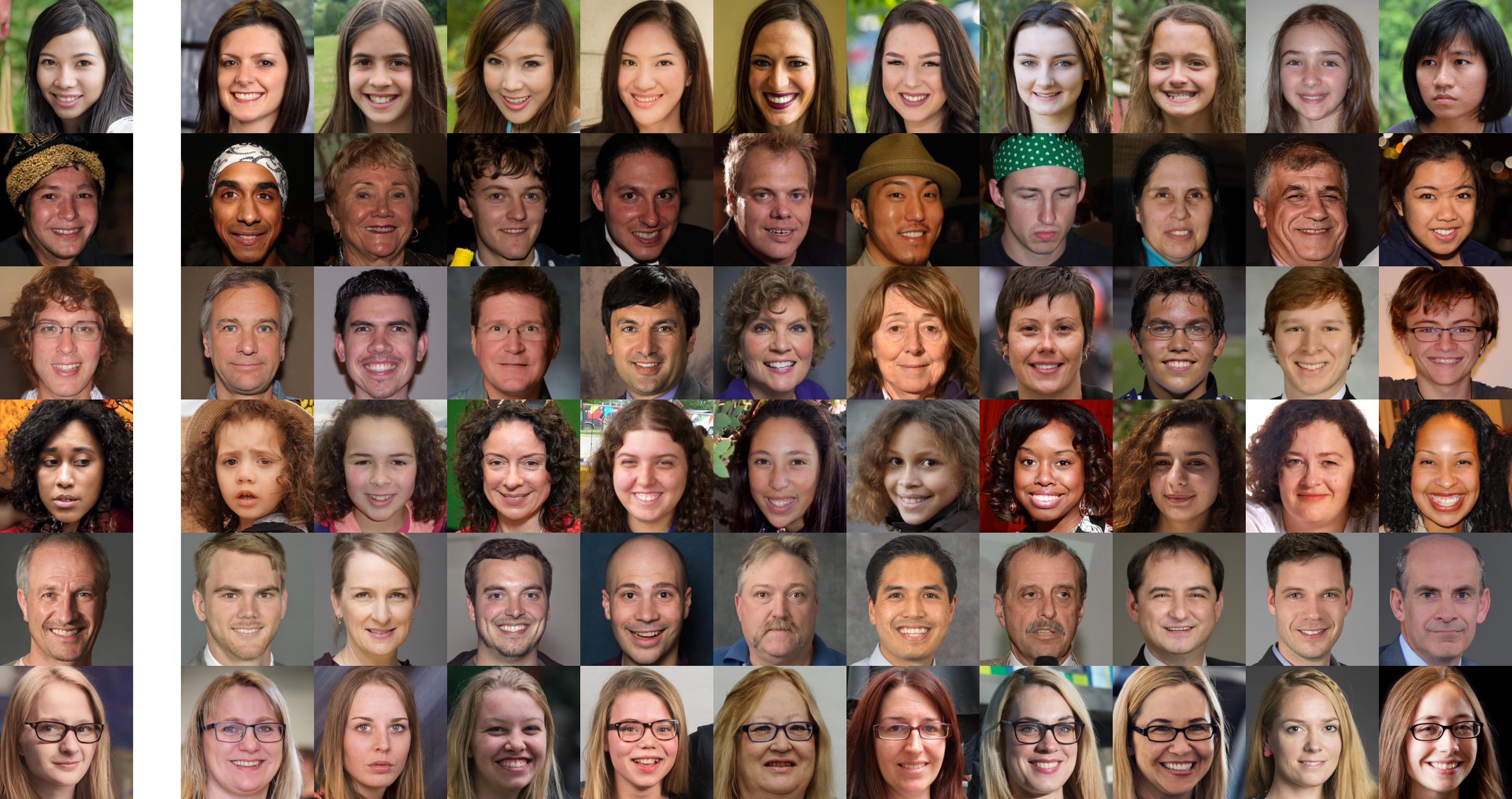}
    \caption{Nearest neighbours for a model trained on FFHQ based on LPIPS distance. The left column contains samples from our model and the right column contains the nearest neighbours in the training set (increasing in distance from left to right).}
\end{figure*}

\begin{figure*}[h]
    \centering
    \includegraphics[width=\textwidth]{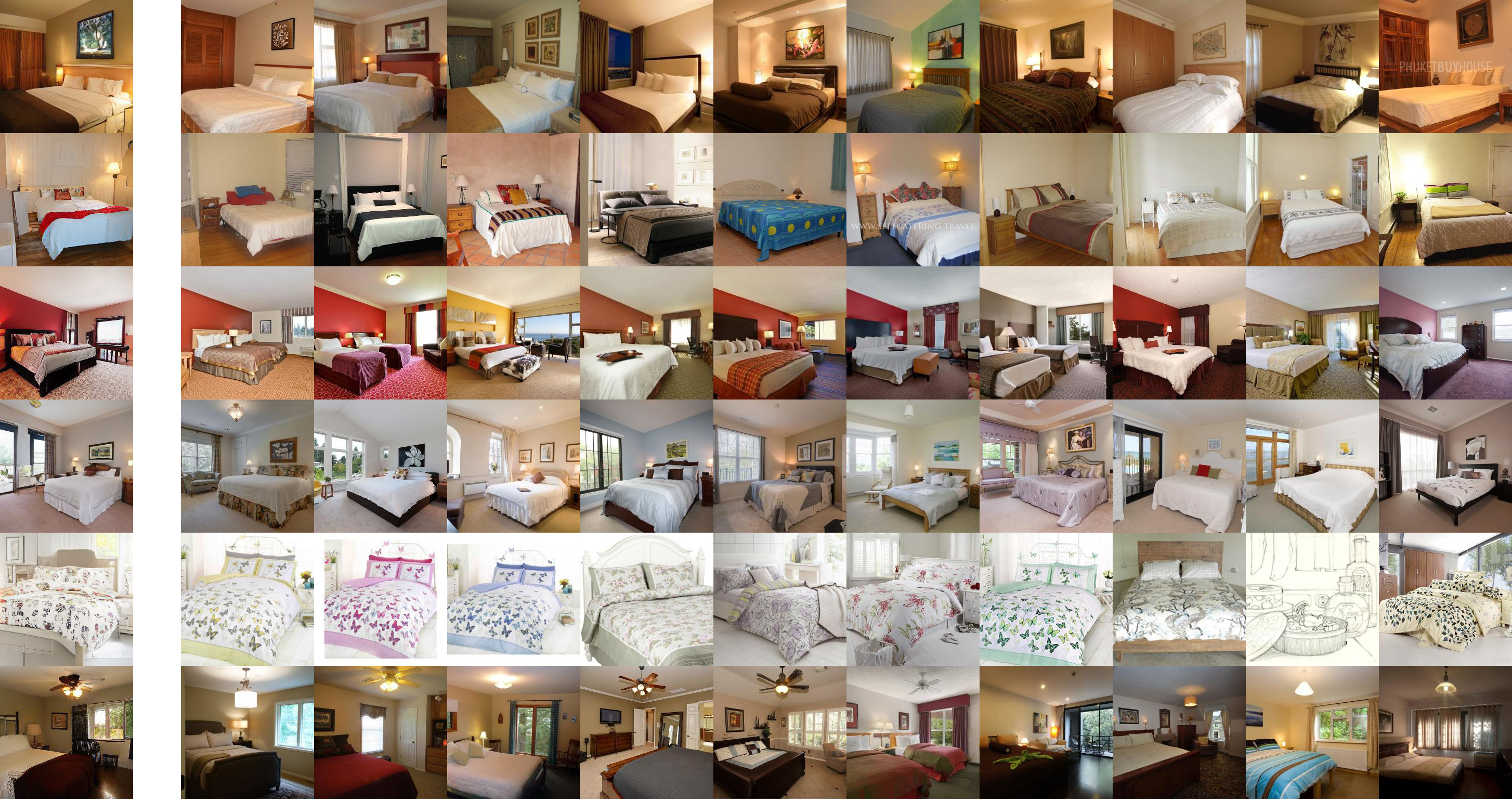}
    \caption{Nearest neighbours for a model trained on LSUN Bedroom based on LPIPS distance. The left column contains samples from our model and the right column contains the nearest neighbours in the training set (increasing in distance from left to right).}
\end{figure*}

\end{document}